\documentclass[letterpaper, 10 pt, conference]{ieeeconf}  % Comment this line out if you need a4paper

\IEEEoverridecommandlockouts                              % This command is only needed if 
\usepackage{graphicx}
\usepackage{amsmath}
\usepackage{amsfonts}
\usepackage{bm}
\usepackage{booktabs}
\usepackage{color}
\usepackage{url}
\usepackage{amsmath}
\usepackage{algorithm}
\usepackage{algpseudocode}
\usepackage{mdframed}
\usepackage{hyperref}
\usepackage{multirow}
\usepackage{comment}

\title{\LARGE \bf
Open-Vocabulary Action Localization with Iterative Visual Prompting
}

\author{Naoki Wake$^{1}$, Atsushi Kanehira$^{1}$, Kazuhiro Sasabuchi$^{1}$, Jun Takamatsu$^{1}$, and Katsushi Ikeuchi$^{1}$% <-this % stops a space
\thanks{$^{1}$Authors are with Applied Robotics Research, Microsoft, Redmond, WA, USA}%
}

\begin{document}

\maketitle
\thispagestyle{empty}
\pagestyle{empty}

%%%%%%%%%%%%%%%%%%%%%%%%%%%%%%%%%%%%%%%%%%%%%%%%%%%%%%%%%%%%%%%%%%%%%%%%%%%%%%%%
\begin{abstract}
Video action localization aims to find the timings of specific actions from a long video. Although existing learning-based approaches have been successful, they require annotating videos, which comes with a considerable labor cost. This paper proposes a training-free, open-vocabulary approach based on emerging off-the-shelf vision-language models (VLMs). The challenge stems from the fact that VLMs are neither designed to process long videos nor tailored for finding actions. We overcome these problems by extending an iterative visual prompting technique. Specifically, we sample video frames and create a concatenated image with frame index labels, allowing a VLM to identify the frames that most likely correspond to the start and end of the action. By iteratively narrowing the sampling window around the selected frames, the estimation gradually converges to more precise temporal boundaries. We demonstrate that this technique yields reasonable performance, achieving results comparable to state-of-the-art zero-shot action localization. These results support the use of VLMs as a practical tool for understanding videos. Sample code is available at \href{https://microsoft.github.io/VLM-Video-Action-Localization/}{https://microsoft.github.io/VLM-Video-Action-Localization/}
\end{abstract}
%%%%%%%%%%%%%%%%%%%%%%%%%%%%%%%%%%%%%%%%%%%%%%%%%%%%%%%%%%%%%%%%%%%%%%%%%%%%%%%%

\section{Introduction}
Video action localization aims to find the timings of specific actions from a long video. Extracting precise timing of an action is a fundamental technique in computer vision and related research domains, including video annotation, video editing, and automatic video collection from unlabeled datasets. For example, in the context of robotics, accurately extracting videos corresponding to specific actions will greatly benefit from analyzing human action in several robot-teaching frameworks where robots learn from human demonstrations, such as Learning from Demonstration (LfD)~\cite{billard2008survey} and Learning from Observation (LfO)~\cite{ikeuchi1992towards, wake2020verbal, wake2020learning}.

Previous mainstream research on video action localization has primarily focused on training models based on labeled datasets. The typical approach is a frame-by-frame method where a pipeline classifies frames or a series of consecutive frames with predefined action labels %for determining which action is being executed at each point in time in the video
~\cite{qing2021temporal,lin2019bmn,lin2018bsn}. While these methods have been successful on several benchmarks, they have limitations in label extensibility. Specifically, these methods require additional data collection and model re-training when new labels are added, which is a bottleneck for the quick and convenient application of these methods to data outside benchmark tasks.
\begin{figure}[t]
  \centering
  \includegraphics[width=\columnwidth]{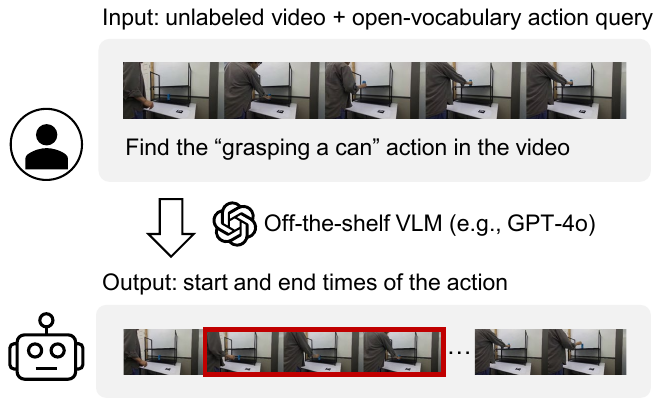}
  \caption{Open-vocabulary video action localization aims to find the start and end times of an action specified by open-vocabulary free-text queries. We propose a training-free approach that leverages an off-the-shelf vision-language model (e.g., OpenAI's GPT-4o).}
  \label{fig:oval}
\end{figure}

\begin{figure*}[tbp]
  \centering
  \includegraphics[width=0.8\textwidth]{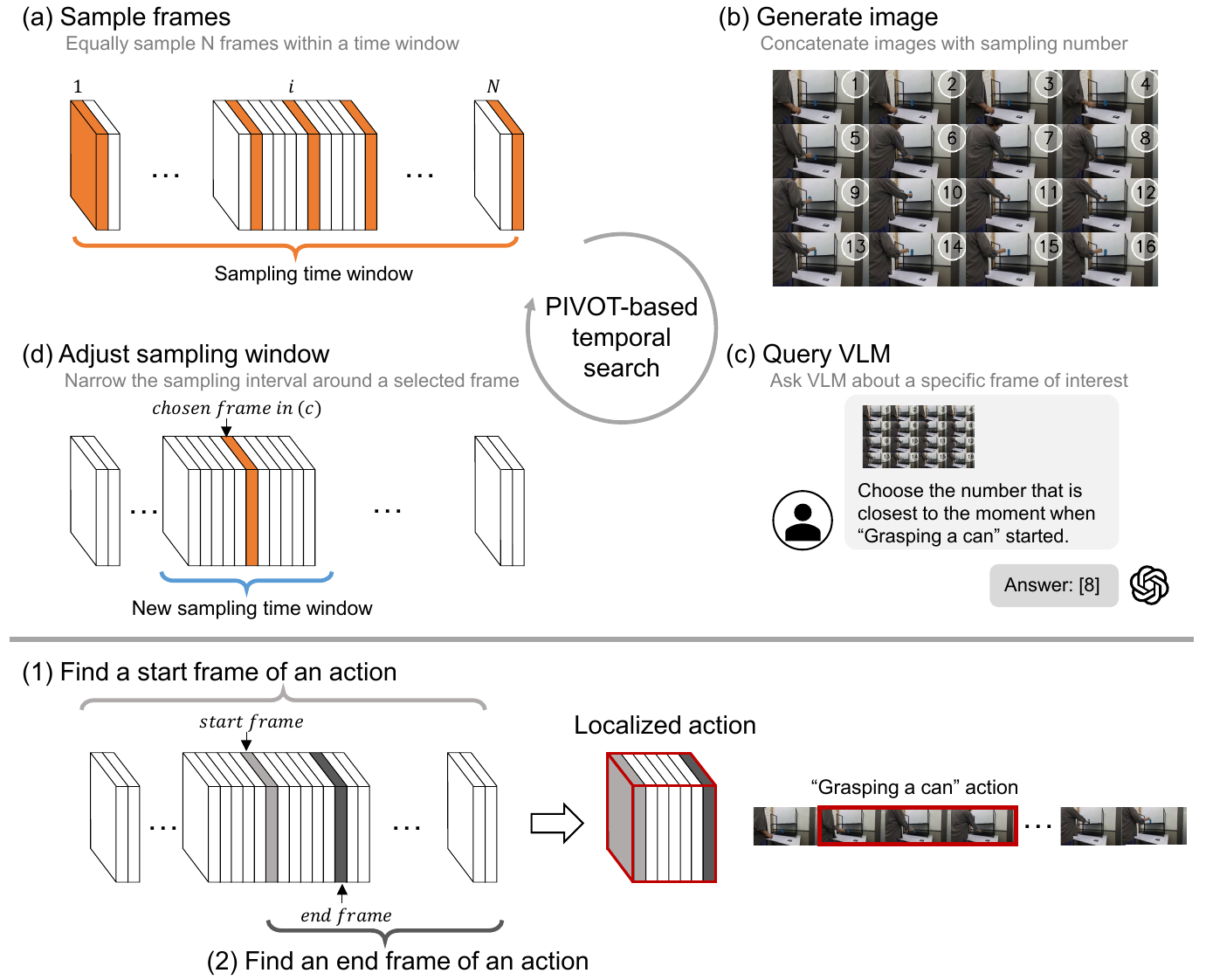}
  \caption{The proposed pipeline for open-vocabulary video action localization using a VLM consists of the following steps: (a) Frames are sampled at regular intervals from a time window, covering the entire video during the first iteration. (b) The sampled frames are then tiled in an image with annotations indicating the time order of the frames. (c) This image is then fed into a VLM to identify the frames closest to a specific timing of an action (e.g., the start time of an action). (d) The sampling window is updated by centering on the selected frame with a narrower sampling interval. Bottom panel (1) For general action localization, the start time of the action in the video is determined by iterating steps (a) to (d). Bottom panel (2) By estimating the end time of the action in the same manner, the action is localized in the video.}
  \label{fig:oval-pipeline}
\end{figure*}
In this paper, we introduce a pipeline for Open-vocabulary Video Action Localization (OVAL), which does not require pre-training and can address action labels in an open-vocabulary form (Fig.~\ref{fig:oval}). Our approach is inspired by a recently proposed method called Prompting with Iterative Visual Optimization (PIVOT)~\cite{nasiriany2024pivot}. The original PIVOT aims to utilize Vision-Language Models (VLMs) to infer an embodied action by framing the problem as iterative visual question answering. In this paper, we propose a simple yet effective approach to apply PIVOT to find temporal segments of actions. Specifically, we sample video frames and create a concatenated image with frame index labels, allowing a VLM to identify the frames that most likely correspond to the start and end of the action. By iteratively narrowing the sampling window around the selected frames, the estimation gradually converges to more precise temporal boundaries, thereby achieving action localization. Although the recent advancements in VLM technology have supported the development of OVAL~\cite{tian2024open,nguyen2024one,wu2024open,hyun2024exploring,gupta2024open,li2024detal}, these methods still rely on domain-specific pre-training, and to the best of our knowledge, there are no training-free methods that utilize off-the-shelf VLMs directly.

Using OpenAI's GPT-4o~\cite{OpenAI} as an example of VLMs, we evaluated the performance of the proposed pipeline on an existing breakfast cooking dataset and our own labeled cooking dataset. We checked how the pipeline works for a typical robot-teaching condition in which transitions between high-level tasks need to be grounded to a human demonstration video~\cite{wake_chatgpt, wake2023gpt, yanaokura2022multimodal} (e.g., grasp, pick up, move, put, and release an object). As a result, we show that the model achieved over 60\% accuracy in terms of mean-over-frame (MoF), outperforming a baseline method. Although the performance appeared to be inferior to existing learning-based methods, the results demonstrated the feasibility of the proposed method. Furthermore, we tested the proposed pipeline for a general action-localization task using a sport video benchmark and showed that it performed comparably to the state-of-the-art zero-shot methods. In addition, to offer further insights into its application, we explored the impact of several visual prompting parameters and present qualitative results for extracting specific actions from an unlabeled dataset. The contributions of this research are:
\begin{itemize}
    \item Proposing a novel pipeline for OVAL by extending PIVOT.
    \item Demonstrating its effectiveness using existing benchmarks and our own benchmarks for robot teaching.
    \item Reporting the impact of parameters of several components and the limitations of this method.
\end{itemize}

The primary advantage of this method is its training-free nature for OVAL, which enables action localization across diverse contexts. We believe its adaptability extends beyond robot teaching, making it applicable to a wide range of research fields and practical use cases.

\section{Related Work}
\subsection{Model-based Action Localization}
Model-based temporal action localization can be broadly categorized into anchor-based methods and boundary-based methods. Anchor-based methods, also known as top-down methods, mainly use multi-scale temporal sliding windows to detect pre-trained actions within untrimmed videos~\cite{buch2019end, heilbron2016fast,escorcia2016daps,gao2017turn,xu2017r,chao2018rethinking,shou2016temporal,huang2019decoupling,zhao2017temporal,zeng2019breaking,ji2019learning,buch2017sst}. In contrast, boundary-based methods, also known as bottom-up methods, initially detect the temporal boundaries of actions, followed by further analysis of the proposal regions estimated from these boundaries~\cite{lin2020fast,lin2019bmn,lin2018bsn,zhao2017temporal,su2021bsn++,liu2019multi}. Boundary-based methods typically involve a two-stage pipeline consisting of temporal proposal generation and action classification~\cite{shou2016temporal,zhao2017temporal,zeng2019breaking,huang2019decoupling}, while end-to-end approaches have also been proposed~\cite{lin2017single,buch2019end,tan2021relaxed,bueno2023leveraging}. The proposed method follows a boundary-based approach, as it aims to detect the start and end boundaries of actions. Additionally, unlike other methods, our VLM-based method accommodates open-vocabulary queries. This allows for the addition of new actions without the need for data collection or re-training.

\subsection{Open-vocabulary Video Action Localization}
OVAL can be broadly categorized into two approaches. The first approach outputs the closest matching action label from a pre-trained action set, while the latter is class-label-free, aiming to localize actions specified by a free-text query. In the former approach, LLMs or high-level scene understanding are used to bridge the gap between the query and pre-trained action labels~\cite{nguyen2024one, tian2024open}. In a typical implementation of the latter approach, actions are first detected and then the corresponding action segment is determined based on the relationship between video features and the query~\cite{li2024detal}. Pre-trained multimodal models are often used to achieve these objectives~\cite{hyun2024exploring,gupta2024open,ju2022prompting,rathod2022open,wu2024open}. The proposed method follows the class-label-free approach, leveraging off-the-shelf models without additional training. Its unique advantage lies in being training-free and using VLMs to directly match action detection with language in a single stage.

Recent studies have also highlighted the significance of open-vocabulary approaches in various other tasks beyond action localization. For instance, open-vocabulary learning has been extensively surveyed in \cite{wu2024towards}, and open-vocabulary methods have been proposed for multiple object tracking \cite{li2023ovtrack}, segmentation \cite{li2024omg}, and video instance segmentation \cite{wang2023towards}, demonstrating a growing interest in more generalized and flexible VLMs. These approaches share a common goal of avoiding the close-set assumption, thereby offering broader applicability and robustness for real-world tasks.

\section{Method}\label{method}
\subsection{Core Pipeline}
Figure~\ref{fig:oval-pipeline} shows the core pipeline of the proposed method, aligning with the original illustration of PIVOT~\cite{nasiriany2024pivot}. The pipeline starts by sampling a number of frames at regular intervals from a time window of the given video (Fig.~\ref{fig:oval-pipeline}(a)). In the first iteration, the sampling window is set to be identical to the length of the video to capture the entire sequence of the video. The number of frames to sample is a hyperparameter and Fig.~\ref{fig:oval-pipeline} illustrates the case of 16 frames. The sampled frames are then tiled in an image with the annotation of the time order of the frames (Fig.~\ref{fig:oval-pipeline}(b)). The image is then fed into a VLM to identify the frame closest to a specific action timing; Fig.~\ref{fig:oval-pipeline}(c) illustrates the start time of an action. Finally, the sampling window is updated by centering on the selected frame with a narrower sampling interval (Fig.~\ref{fig:oval-pipeline}(d)). Note that due to the sparsity of sampling, the exact moment of an action may not be present in the query image during the first iteration. However, the method can still identify the relevant moment precisely by iteratively narrowing the sampling window around the most likely frame until a specific temporal resolution is reached. Hereinafter, we will refer to this method as Temporal PIVOT (T-PIVOT).

For general action localization, we first estimate the start time of the action in the video using T-PIVOT (Fig.~\ref{fig:oval-pipeline} bottom panel(1)). Then, by resetting a sampling window for the period after the start time, we estimate the end time of the action in the same manner (Fig.~\ref{fig:oval-pipeline} bottom panel(2)). By using the obtained start and end times, we localize the action of interest. Importantly, since this method involves querying the VLM, it allows us to specify actions in open-vocabulary free-text queries.

This pipeline can be extended to address several sub-scenarios, including but not limited to: finding the transitions between a series of actions within a video, detecting a specific event in a video, and collecting multiple instances of an action for the purpose of data curation. 

\subsection{Implementation of T-PIVOT}
Algorithm~\ref{algo:frameselection} shows the specific pipeline we implemented for the experiment in this paper. The computation yields the start and end times of a set of tasks in a video by applying T-PIVOT in parallel. We used GPT-4o for all the experiments. The sampling window was halved with each iteration, and this process was repeated for a specific number of iterations.

\begin{algorithm}
\caption{Parallel T-PIVOT}
\label{algo:frameselection}
\begin{algorithmic}[1]
\small
\Require $video$, $tasks$
\State $results \gets \{\}$
\State $center\_time \gets video\_length / 2$
\State $window \gets video\_length$
\ForAll{$task \in tasks$} \textbf{in parallel}
    \For{$i = 1$ to $iterations$}
        \State $image \gets SampleFrames(center\_time, window)$
        \State $selected\_time \gets QueryVLM(image, task)$
        \State $center\_time \gets selected\_time$
        \State $window \gets window / 2$
    \EndFor
    \State $results[task] \gets center\_time$
\EndFor
\State \Return $results$
\end{algorithmic}
\end{algorithm}

\begin{itemize}
    \item $SampleFrames$ samples a set of frames centered at $center\_time$ within the given $window$, arranges them into a grid image, and returns the resulting image.
    \item $QueryVLM$ uses a VLM to analyze the image and returns the $selected\_time$ corresponding to the frame that most likely represents the target moment of a given $task$.
\end{itemize}

\subsection{Qualitative Results}\label{qualitative_test}
\begin{figure}[ht]
  \centering
  \includegraphics[width=\columnwidth]{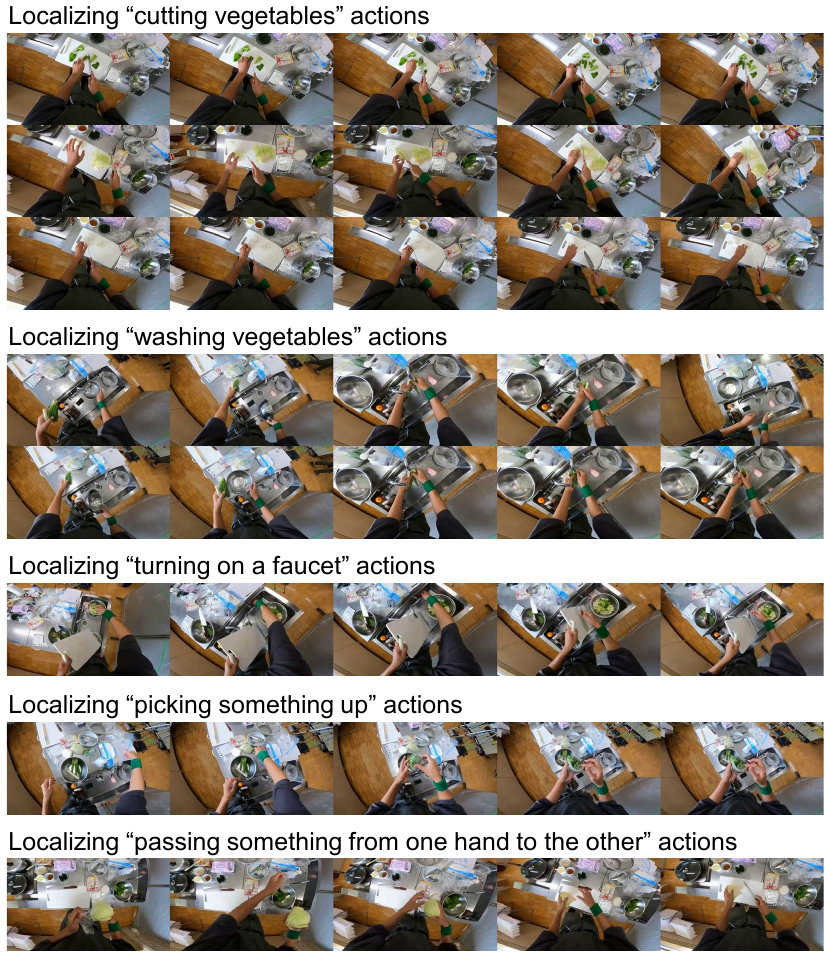}
  \caption{Qualitative results of open-vocabulary video action localization on a cooking video.}
  \label{fig:qualitative_results}
\end{figure}

We qualitatively checked our proposed pipeline using a cooking video that we recorded in-house (Fig.~\ref{fig:oval-pipeline}). This 10-minute first-person video included actions such as taking out, washing, and cutting vegetables. 
To maintain sufficient temporal resolution, we divided the video into five-second intervals and applied Algorithm~\ref{algo:frameselection} to each interval. If an action was detected, the subsequent window was adjusted to begin at the estimated end time of the localized action; otherwise, the window advanced by another five seconds. 
Figure~\ref{fig:qualitative_results} shows the examples of video segments identified for specified actions, demonstrating that the method produced reasonable results.

Notably, T-PIVOT worked not only for static actions that are closely tied to stationary scene information but also for dynamic hand movements. For instance, actions such as ``cutting vegetables,'' ``washing vegetables,'' and ``turning on a faucet'' could be recognized from a single frame due to the presence of static objects like a cutting board or faucet. In contrast, actions such as ``picking something up'' or ``passing something from one hand to the other'' would not rely on such contextual cues. Instead, they require attention to dynamic hand motion for accurate localization, which cannot be achieved solely by single-frame analysis. Nevertheless, our method successfully localized these actions based on textual descriptions. This success is likely because the input image, formatted by arranging time-series frames, conveyed temporal information such as hand motions to the VLM.

Another notable implication is that T-PIVOT localized second-scale actions within a long, 10-minute video. In cases where the target action occupies only a small portion of the video, the initial iteration of T-PIVOT is unlikely to extract the relevant frames, thus VLMs will not select the frames that include the action. However, by defining a reasonable temporal window (i.e., five seconds) and applying T-PIVOT sequentially, T-PIVOT effectively extracts actions regardless of the original video length. This temporal-window strategy was also employed in the experiment in zero-shot action localization in Sec.~\ref{comparison}.

\section{Experiment}
We evaluated the performance of the proposed pipeline in light of practical use cases. Specifically, the pipeline was evaluated in the context of a typical robot-teaching condition on top of task and motion planning (TAMP)~\cite{garrett2021integrated}, where a system needs to ground high-level task plans (e.g., grasp, pick up, move, put, and release an object) to a human demonstration video~\cite{wake_chatgpt, wake2023gpt, yanaokura2022multimodal}. To this end, we used an existing benchmark dataset and our own labeled cooking dataset. We focused on cooking actions because they reflect realistic scenarios and pose challenges due to the presence of cluttered objects in the scene.

\subsection{Benchmark}
\noindent \textbf{Breakfast Dataset}~\cite{kuehne2014language} consists of third-person videos containing ten activities related to breakfast preparation, such as pouring a drink and making a sandwich. Each video is annotated without gaps along the timeline with more detailed action labels, such as taking a cup and stirring. The videos contain 52 different individuals across 18 different kitchens, totaling 1712 videos. Since we are proposing a training-free approach, we used only the testing dataset of 335 videos based on the split used in~\cite{hussein2019timeception}.

\noindent \textbf{Fine-grained Breakfast Dataset} consists of first-person videos of human cooking activities with fine-grained robotics tasks. Existing datasets of human actions, including the Breakfast Dataset~\cite{kuehne2014language}, are labeled with coarse categories and simple motion labels often consist of multiple robotic manipulations. For example, a motion labeled as ``pour milk'' needs a sequential robotic manipulation of moving a robot hand towards a milk carton, grasping it, picking it up, moving it near a cup, rotating and holding the milk carton, adjusting its pose, moving back to the table, and releasing it~\cite{goyal2017something, damen2020epic}---there is a huge gap between the granularity of action in the computer vision community and the robotics community. To our knowledge, however, no existing datasets label tasks based on fine-grained robotic actions. To gain insights into the usefulness of the proposed pipeline for robot teaching, we prepared a Fine-grained Breakfast Dataset by manually annotating an existing dataset of breakfast preparation~\cite{saudabayev2018human} using a third-party annotation tool~\cite{video-annotator}.

Based on the Kuhn-Tucker theory~\cite{kuhn1956related} explaining solution space of linear simultaneous inequalities, we defined tasks as transitions of constraints imposed on an object~\cite{ikeuchi2024semantic, ikeuchi2024applying}. As a result, we obtained a task set that is necessary and sufficient to explain manipulation, such as picking up, moving, placing, and rotating (refer to~\cite{ikeuchi2024semantic} for the complete list of tasks). 113 videos were used in this study.

\subsection{Task Definition}
Given a video $V$ and a sequence of task labels $L_i$ ($i = 1 \ldots N$) contained within it, our goal is to estimate the timing of these transitions $T_{i \rightarrow i+1}$ ($i = 1 \ldots N-1$). We assume that the task sequence is continuous. As a baseline method, we partition the video uniformly based on the length of the sequence, allocating equal time intervals to each task.
We evaluate our method using the following metrics commonly used in temporal action localization~\cite{bueno2023leveraging}:
\begin{itemize}
    \item \textbf{MoF (Mean-over-Frames)} represents the percentage of correctly predicted frames out of the total number of frames in a video.
    \item \textbf{IoU (Intersection over Union)}:  measures the overlap between the predicted and ground truth segments for each action class. It is calculated as the ratio of the intersection over the union of the two sets.
    \item \textbf{F1-score}: represents the harmonic mean of precision and recall, providing a balanced measure of prediction accuracy.
\end{itemize}

\subsection{T-PIVOT for Recognizing Seamless Action Transitions}\label{TaskTransitionEstimation}
By applying the T-PIVOT method described in Sec.~\ref{method}, we estimated the timing of task transitions within the video as follows. First, the start time of each task, $T_{i\_start}$ ($i = 1 \ldots N$), was estimated using the T-PIVOT in parallel processing. The end time of each task, $T_{i\_end}$ ($i = 1 \ldots N$), was similarly estimated. This approach yielded an estimated series of start and end times for each task. In this experiment, we assumed that tasks were seamlessly connected throughout the video. Therefore, the end time of one task was identical to the start time of the subsequent task. To ensure this, we calculated this time as the midpoint between the estimated end time of the current task and the estimated start time of the next task:
\begin{equation}
    T_{i \rightarrow i+1} = (T_{i\_end} + T_{i+1\_start})/2.
\end{equation}
Specifically, the transition time $T_{i \rightarrow i+1}$ ($i = 1 \ldots N-1$) was calculated using Algorithm~\ref{algo:seamlessframeselection}.
\begin{algorithm}
\caption{Task Transition Estimation}
\label{algo:seamlessframeselection}
\begin{algorithmic}[1]
\small
\Require $video$, $tasks$
\State $T_{i\_start} \gets EstimateStartTimes(video, tasks)$
\For{$i = 1$ to $N-1$}
    \If{$T_{i+1\_start} < T_{i\_start}$}
        \State $T_{i+1\_start} \gets T_{i\_start}$
    \EndIf
\EndFor

\State $T_{i\_end} \gets EstimateEndTimes(video, tasks)$
\For{$i = 1$ to $N-1$}
    \If{$T_{i\_end} > T_{i+1\_end}$}
        \State $T_{i\_end} \gets T_{i+1\_end}$
    \EndIf
\EndFor

\For{$i = 1$ to $N-1$}
    \State $T_{i \rightarrow i+1} \gets (T_{i\_end} + T_{i+1\_start}) / 2$
\EndFor

\State \Return $T_{i \rightarrow i+1}$
\end{algorithmic}
\begin{tabular}[t]{p{0.45\textwidth}}
   $EstimateStartTimes$ and $EstimateEndTimes$ correspond to Algorithm~\ref{algo:frameselection}, which aims to find the start and end times, respectively.
\end{tabular}
\end{algorithm}

\subsection{Action-order Aware Prompting}
To estimate the start and end times of each task in parallel with Algorithm~\ref{algo:seamlessframeselection}, each T-PIVOT process needs to know which task in the task sequence it is focusing on. Without this knowledge, the process cannot determine which instance to focus on; for example, if there are two instances of ``picking up'' in a video, it is not clear which ``picking up'' action is of interest when given a query of ``Please find the picking up action in this video.'' To address this issue, we designed the following prompt using a sequence of task labels $L_i$ ($i = 1 \ldots N$):\\
\textit{``I will show an image sequence of human operation. It contains the following tasks: \{\textbf{task\_sequence}\}. I have annotated the images with numbered circles. Choose the number that is closest to the moment when the (\{\textbf{task\_focus}\}) has started. You are a five-time world champion in this game. Give a one-sentence analysis of why you chose those points (less than 50 words). Provide your answer at the end in a JSON file in this format: \{``points'': []\}.''} \\
Here, \textit{\{task\_sequence\}} lists the sequence of tasks in the format ``1. $L_1$, 2. $L_2$, ...'' and \textit{\{task\_focus\}} refers to the specific task being searched for in this instance (i.e., ``k. $L_k$''). This prompt is designed to make the VLM aware of both the task order and which task is currently being targeted.

\subsection{Performance Across Different Prompting Styles}\label{prompting_style}
We investigated the effect of different prompting methods on the performance of the T-PIVOT. The aim of this experiment is to report the impact of several adjustable parameters to provide insights that may assist readers in designing their own prompts. It is important to note that there is no standard approach for creating prompt images or the method of updating sampling time windows, and the optimal methods may vary depending on the VLM models used and the types of videos analyzed. Therefore, the findings may not necessarily generalize across all models and video types. 

We first investigated the effectiveness of the iterative process on performance. Table~\ref{tab:iter_comparison} shows the impact on performance of varying the number of iterations, using the Breakfast Dataset. In this experiment, we used a 2x2 grid configuration, which led to coarse temporal resolution per iteration. The results showed that overall performance improved over the iterations, suggesting that the iterative process effectively enhances sampling resolution. Notably, even with the coarse 2x2 grid setting, performance peaked within four iterations. For subsequent experiments, the number of iterations was set to four as a practical choice.

Second, we investigated how to create images for inclusion in the prompt, specifically focusing on how many images from a time window should be presented and in what manner. Table~\ref{tab:grid_comparison} and Table~\ref{tab:grid_comparison_arr} show the impact of grid size, which reflects the number of sampled images per iteration, on the results. The highest performance was achieved with a 5x5 grid (i.e., extracting 25 frames from a sampling time window) across the three metrics for the Breakfast Dataset, and 3x3- or 4x4-grid settings for the Fine-grained Breakfast Dataset. These results appear to highlight the trade-off between temporal and spatial resolution in this method---a smaller grid size reduces the temporal resolution in recognition as it limits the number of frames that can be extracted from the video, while a larger grid size lowers the spatial resolution of each frame, under the assumption that the number of iterations and the resolution of the prompt image remain the same.

\begin{figure}[ht]
  \centering
  \includegraphics[width=\columnwidth]{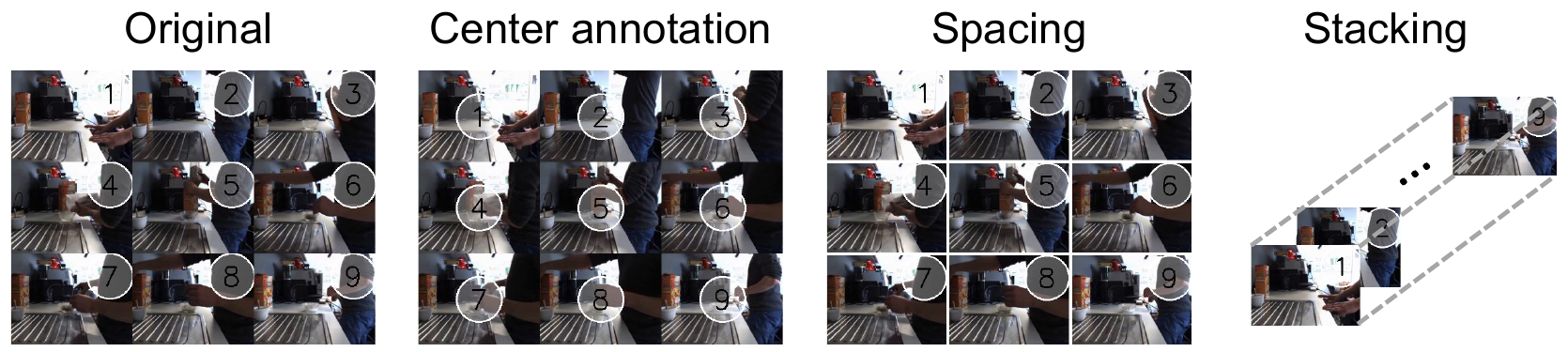}
  \caption{Different types of visual prompting styles tested in this study.}
  \label{fig:visual_parameters}
\end{figure}

Lastly, we investigated other parameters for visual prompting, including the rendering position of the annotation numbers and the tiling spacing (Fig.~\ref{fig:visual_parameters}). We also explored an approach where the images were input into the model as stacked buffers rather than tiled. The highest performance across the three metrics was achieved using either the original tiling method or the stacking method. The inferior performance observed with the center rendering of the annotation numbers might be due to the rendered numbers obscuring important visual content, making it harder to interpret the underlying image.
\begin{table}[ht]
\caption{Performance Metrics Across Iteration Counts (Breakfast Dataset)}
\centering
\renewcommand{\arraystretch}{1.3} % Adjust row height
\begin{tabular}{lccccc}
\hline
\textbf{Iteration count} & \textbf{1} & \textbf{2} & \textbf{3} & \textbf{4} & \textbf{5}\\ 
\hline
MoF & 32.4 & 39.7 & 39.5 & \textbf{40.7} & 38.6 \\
IoU & 18.4 & 23.8 & 24.0 & \textbf{24.5} & 23.2 \\
F1  & 24.4 & 31.6 & 31.9 & \textbf{32.3} & 30.7 \\
\hline
\end{tabular}
\label{tab:iter_comparison}
\footnotesize
\begin{tabular}[t]{p{0.45\textwidth}}
\centering
Images were rendered with 2x2 grid tiling.
\end{tabular}
\end{table}

\begin{table}[ht]
\caption{Performance Metrics Across Grid Sizes (Breakfast Dataset)}
\centering
\renewcommand{\arraystretch}{1.3} % Adjust row height
\begin{tabular}{lcccccc}
\hline
\textbf{Grid size} & \textbf{2x2} & \textbf{3x3} & \textbf{4x4} & \textbf{5x5} & \textbf{6x6} & \textbf{Baseline} \\
\hline
MoF & 40.7 & 52.2 & 59.8 & \textbf{60.1} & 58.0 & 36.0 \\
IoU & 24.5 & 33.2 & 40.3 & \textbf{40.8} & 39.1 & 22.0 \\
F1  & 32.3 & 42.7 & 50.6 & \textbf{51.2} & 49.4 & 30.3 \\
\hline
\end{tabular}
\label{tab:grid_comparison}
\end{table}

\begin{table}[ht]
\caption{Performance Metrics Across Grid Sizes (Fine-grained Breakfast Dataset)}
\centering
\renewcommand{\arraystretch}{1.3} % Adjust row height
\begin{tabular}{lcccccc}
\hline
\textbf{Grid size} & \textbf{2x2} & \textbf{3x3} & \textbf{4x4} & \textbf{5x5} & \textbf{6x6} & \textbf{Baseline} \\
\hline
MoF & 54.9 & \textbf{67.0} & 66.9 & 64.6 & 61.7 & 62.7 \\
IoU & 30.2 & 50.7 & \textbf{52.0} & 50.3 & 45.8 & 49.1 \\
F1  & 38.5 & 61.0 & \textbf{62.7} & 61.5 & 56.7 & 60.6 \\
\hline
\end{tabular}
\label{tab:grid_comparison_arr}
\end{table}

\begin{table}[t]
\caption{Performance Metrics Across Rendering Styles (Breakfast Dataset)}
\centering
\renewcommand{\arraystretch}{1.3} % Adjust row height
\begin{tabular}{lcccc}
\hline
\textbf{Method} & \textbf{Original} & \textbf{Center annotation} & \textbf{Spacing} & \textbf{Stacking}\\
\hline
MoF & \textbf{60.1} & 54.1 & 58.0 & 59.6 \\
IoU & 40.8 & 37.8 & 40.0 & \textbf{45.4} \\
F1  & 51.2 & 48.1 & 50.5 & \textbf{54.8} \\
\hline
\end{tabular}
\label{tab:breakfast_dataset_prompting}
\footnotesize
\begin{tabular}[t]{p{0.45\textwidth}}
\centering
Images were rendered with 5x5 grid tiling.
\end{tabular}
\end{table}

\begin{figure}[ht]
  \centering
  \includegraphics[width=\columnwidth]{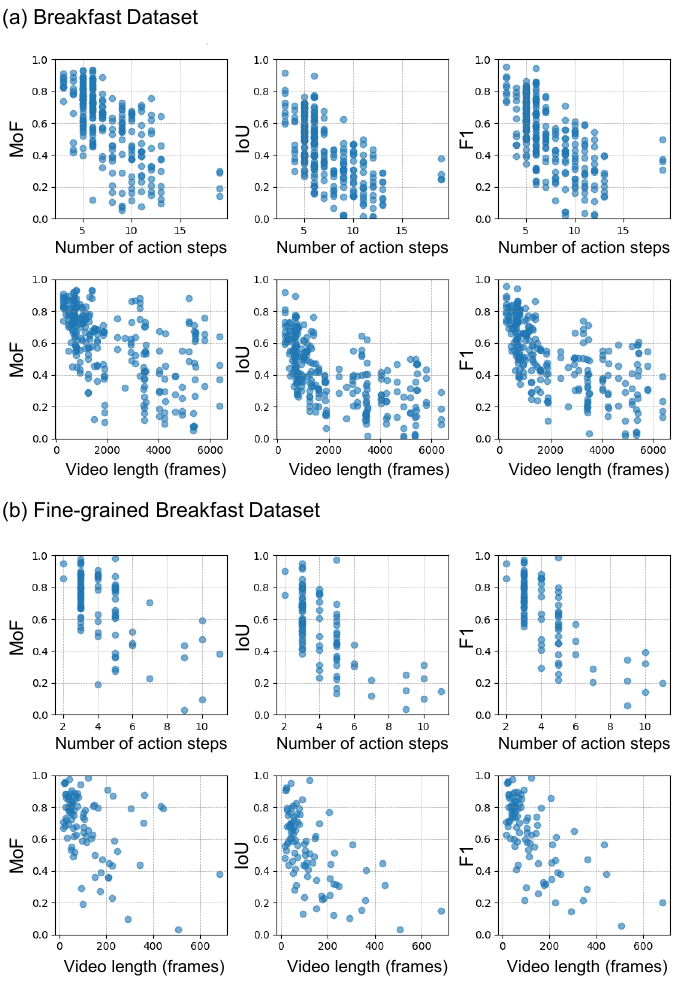}
  \caption{Action localization performance plotted against the number of action steps and the video length. (a) Breakfast Dataset and (b) Fine-grained Breakfast Dataset.}
  \label{fig:visual_parameters_detail}
\end{figure}

Furthermore, to gain deeper insights into the performance of action recognition, we examined the impact of the number of actions in each video and the duration of each video on performance (Fig.~\ref{fig:visual_parameters_detail}). These results indicate that the number of action steps significantly affects the outcomes, highlighting the remaining challenges in recognizing actions in longer videos.

\subsection{Comparison with Existing Methods}\label{comparison}
Table~\ref{tab:comparison_existing} compares our proposed method with existing methods on the Breakfast Dataset, with the reported numbers obtained from~\cite{bueno2023leveraging}. It is important to note that this comparison is not direct due to fundamental differences in experimental assumptions. T-PIVOT assumes the action sequence in the video is given for task-transition estimation. In contrast, learning-based methods typically operate without this information, making the problem setting more challenging for them. Despite these differences, existing methods still outperform our approach, highlighting its limitations compared to the state-of-the-art learning-based techniques.

To gain insights into the zero-shot capabilities of our method, we compared its performance with existing zero-shot action localization methods. To this end, we used the THUMOS14 Dataset and its officially provided evaluation code~\cite{THUMOS14}. Importantly, this task is not transition estimation because the THUMOS14 Dataset does not assume that tasks are continuously present in the video. Thus, we employed a different approach from the one in Sec.~\ref{TaskTransitionEstimation}. Similar to Sec.~\ref{qualitative_test}, we divided each video into fixed time intervals and applied Algorithm~\ref{algo:frameselection} to each interval to localize the start and end of actions. Specifically, the video was divided into intervals of five seconds, and a 5x5 grid tiling with four iterations was applied. To operate T-PIVOT, we assumed that the action labels contained in the video were known. Following the methods in~\cite{li2024detal}, we report the mean average precision (mAP), which represents the mean of the average precision values computed at different IoU thresholds (i.e., [0.3, 0.4, 0.5, 0.6, 0.7]). Table~\ref{tab:comparison_existing_openvocab} shows the comparison results, where the reported numbers were obtained from~\cite{li2024detal}. These results show that the proposed method achieves state-of-the-art performance. Importantly, the existing method employs pre-training even though the performance is tested under zero-shot conditions, while our method does not require any pre-training.

\begin{table}[t]
    \caption{Comparison with existing action localization methods (Breakfast Dataset)}
    \centering
    \renewcommand{\arraystretch}{1.2} % 行間を広げる
    \begin{tabular}{l|c c c}
    \toprule
    % \multicolumn{4}{c}{\textbf{Breakfast Dataset}} \\
    % \midrule
    Methods & \textbf{MoF} & \textbf{IoU} & \textbf{F1} \\
    \midrule
    LSTM+AL~\cite{aakur2019perceptual} & 42.9 & 46.9 & - \\
    VTE~\cite{vidalmata2021joint} & 52.2 & - & - \\
    DGE~\cite{dimiccoli2020learning} & 59.5 & \underline{48.5} & 51.7 \\
    TW-FINCH~\cite{sarfraz2021temporally} & 62.7 & 42.3 & 49.8 \\
    ABD~\cite{zexing2022} & \underline{64.0} & - & \underline{52.3} \\
    Triplet~\cite{bueno2023leveraging} & \textbf{65.1} & \textbf{52.1} & \textbf{54.6} \\
    \midrule
    T-PIVOT & 60.1 & 40.8 & 51.2 \\
    % T-PIVOT (Phi-3-vision) & 33.7 & 20.3 & 27.6 \\
    % T-PIVOT (Phi-3-vision) & 31.9 & 18.5 & 25.4 \\
    \bottomrule
    \end{tabular}
    \footnotesize
    \begin{tabular}[t]{p{0.45\textwidth}}
    The top and second-best scores are bolded and underlined, respectively.
    \end{tabular}
    \label{tab:comparison_existing}
\end{table}
\begin{table}[t]
    \centering
    \caption{Comparison with existing zero-shot methods (THUMOS14 Dataset)}
    \resizebox{0.48\textwidth}{!}{
    \begin{tabular}{l|c c c c c |c}
        \toprule
        \textbf{Methods} & 0.3 & 0.4 & 0.5 & 0.6 & 0.7 & \textbf{Avg mAP} \\
        \midrule
        TMaxer~\cite{tang2023temporalmaxer} & 10.6 & 9.4 & 8.0 & 6.2 & 4.4 & 7.7 \\
        ActionFormer~\cite{zhang2022actionformer} & 11.3 & 10.0 & 8.4 & 6.6 & 4.6 & 8.2 \\
        TriDet~\cite{shi2023tridet} & 15.2 & 13.2 & 10.8 & 7.9 & 5.2 & 10.5 \\
        B-II~\cite{radford2021learning} & 21.0 & 16.4 & 11.2 & 6.3 & 3.2 & 11.6 \\
        B-I~\cite{radford2021learning} & 27.2 & 21.3 & 15.3 & 9.7 & 4.8 & 15.7 \\
        Eff-Prompt~\cite{ju2022prompting} & 37.2 & 29.6 & 21.6 & 14.0 & 7.2 & 21.9 \\
        STALE~\cite{nag2022zero} & \underline{38.3} & 30.7 & 21.2 & 13.8 & 7.0 & 22.2 \\
        DeTAL~\cite{li2024detal} & \underline{38.3} & \underline{32.3} & \textbf{24.4} & \textbf{16.3} & \underline{9.0} & 24.1 \\
        OVTAL~\cite{gupta2024open} & - & - & - &  -& - & \underline{24.9} \\
        \midrule
        T-PIVOT & \textbf{45.4} & \textbf{33.6} & \underline{23.2} & \underline{15.5} & \textbf{11.4} & \textbf{25.8}\\
        \bottomrule
    \end{tabular}
    }
    \footnotesize
    \begin{tabular}[t]{p{0.45\textwidth}}
    The top and second-best scores are bolded and underlined, respectively.
    \end{tabular}
    \label{tab:comparison_existing_openvocab}
\end{table}

\section{Discussion and Limitations}
In this paper, we proposed a method for VLM-empowered OVAL, using GPT-4o as an example of VLMs. Despite its straightforward iterative pipeline, our method demonstrated reasonable performance in both third-person and first-person videos. The results were promising in task-transition estimation, although limitations remain—for example, performance tends to degrade on videos with longer action sequences. Additionally, the proposed method showed competitive results among existing zero-shot methods. Notably, the unique advantage of the proposed method lies in its training-free nature---while the prior methods rely on pre-training even under zero-shot conditions, our method requires no model training nor data collection, allowing for immediate and flexible application to new vocabularies and dataset domains. Such adaptability makes it suitable for various applications such as video annotation and editing.
\subsection{Limitation}
The T-PIVOT method involves several hyperparameters in visual prompting style, including language prompts, grid size, and the number of iterations, which may require tuning for each use case. Nevertheless, the experiments on different prompting styles in Sec.~\ref{prompting_style} have revealed the general trends. First, the iteration number tends to improve performance up to a certain point. Next, the optimal number of frames for sampling depends on a trade-off between spatial and temporal resolution---a smaller grid size reduces temporal resolution by limiting the number of extracted frames, whereas a larger grid size lowers spatial resolution per frame, under the assumption that the number of iterations and the prompt image resolution remain the same. Finally, the method's performance is impacted by the rendering position of the annotation numbers, suggesting the importance of rendering that avoids interference with image clarity. These results provide practical guidance for designing visual prompting.

As another limitation, the performance of T-PIVOT can also be influenced by the choice of VLM. To further examine its impact, we conducted an additional experiment using Phi-3-vision~\cite{phi3vision}, one of the state-of-the-art open-source VLMs, applying it to the Breakfast Dataset under the same conditions as GPT-4o. Phi-3-vision showed considerably lower performance (MoF: 31.9, IoU: 18.5, and F1: 25.4) compared to GPT-4o (MoF: 60.1, IoU: 40.8, and F1: 51.2; see Table~\ref{tab:comparison_existing}). While open-source VLMs generally perform well in describing single images~\cite{abdin2024phi}, T-PIVOT requires extracting temporal information from a single image that summarizes an image sequence. Furthermore, this task demands interpretation of textual prompts to identify the start and end points of a specified action, making it more complex than standard image description. These results indicate that our approach benefits from models with strong video reasoning capabilities. 

Additionally, since our method requires multiple queries to the VLM, it tends to take more time compared to typical learning-based methods. This increased latency is a trade-off for the flexibility and adaptability provided by open-vocabulary models, which do not require task-specific training. While this drawback may limit some real-time applications, our method remains feasible for offline analysis or tasks where flexibility in specifying target actions is prioritized.

While our method involves hyperparameter tuning and relies on VLM performance, its novelty lies in the use of off-the-shelf VLMs for OVAL. Our primary contribution is bridging the multimodal perception capabilities of pre-trained VLMs with action localization through an iterative prompting strategy.

\section{Conclusion}
This paper introduced a novel approach for OVAL leveraging the PIVOT methodology, which eliminates the need for data collection and model training. Testing with benchmark datasets has demonstrated that the proposed method offers promising zero-shot capabilities without the need for model training. Considering several limitations of this method, future work will focus on refining visual prompting techniques and optimizing the system to enhance its applicability across a broader range of video analysis tasks, including automated video labeling and advanced robotics training.

\section*{Acknowledgment}
We thank Dr. Sakiko Yamamoto, Dr. Etsuko Saito (Ochanomizu University), and Dr. Midori Otake (Tokyo Gakugei University) for their help in annotating the Fine-grained Breakfast Dataset. This study was conceptualized, conducted, and written by the authors, and an AI, OpenAI's GPT-4o, was used for proofreading.
\bibliographystyle{IEEEtran}
\bibliography{bib}
\end{document}